\documentclass[11pt, a4paper,  notitlepage]{extarticle}

\usepackage[left=3.5cm, right=2.5cm, top=3cm, bottom=3cm, bindingoffset=0cm]{geometry}
\usepackage{graphicx}
\usepackage{mathtools}
\usepackage[affil-it]{authblk}
\usepackage{url}
\usepackage{amsmath}
\usepackage{amssymb}
\usepackage{multicol}
\usepackage{float}
\usepackage[section]{placeins}
\usepackage{hyperref}

\begin{document}
\sloppy
\title{Financial News Analytics  Using Fine-Tuned Llama 2 GPT Model}
\author{Bohdan M. Pavlyshenko \\  
\small{ email:b.pavlyshenko@gmail.com, LinkedIn:https://www.linkedin.com/in/bpavlyshenko/}}
\maketitle


\begin{abstract}
The paper considers the possibility to fine-tune Llama 2 GPT large language model (LLM) for the multitask analysis of financial news. For fine-tuning, the PEFT/LoRA based approach was used. In the study, 
the model was fine-tuned for the following tasks: analysing a text from financial market perspectives, 
highlighting main points of a text, summarizing a text and extracting named entities with appropriate sentiments. 
The obtained results show that the fine-tuned Llama 2 model can perform a multitask financial news analysis with a specified structure of response, part of response can be a structured text and another part of data can have JSON format  for further processing. Extracted sentiments for named entities can be considered as 
predictive features in supervised machine learning models with quantitative target variables. 

financial news.
\end{abstract}

\section{Introduction}
Large language models (LLM), based on generative pre-trained transformers (GPT),  such as ChatGPT  show high efficiency in the analysis of complex texts. 
These days, we can observe the emerging of many new smaller open source LLMs, e.g. Llama, Falcon, GPT4All, GPT-J, etc. 
Open source LLMs can be fine-tuned for specific custom problems and deployed on custom servers, e.g. in cloud computing services such as AWS, GCP. 
 LLMs have some new features as compared to conventional language models based on transformers. One of them is zero-shot and few-shot learning, which 
 consists in good performance of the model when we show it only few training examples or even no examples at all, but only the instructions describing what should be done. Another important feature is the reasoning when a model can generate new patterns and conclusions which are based on an input prompt and facts known by the model and which were not included into it directly during a training process. So, the model can generate analytical texts with unexpected but useful chains of thoughts.  One of the approaches of using LLMs is based on retrieval augmented generation (RAG), which uses the results from other services e.g. relational database, semantic search, graph database in the input prompt for LLM. In this case,   the response  can be treated as the combination of  external results and LLM knowledge. This approach makes it possible to use the standard LLM from the box without fine-tuning it for processing custom data. It processes each custom document separately, but it does not allow us to make reasoning using all custom knowledge base. Taking into account all patterns in custom data, we need to fine-tune the model on custom texts. 

In predictive analytics, it is important to take into account informational trends from different data sources, including news websites and social networks. 
  The work~\cite{pavlyshenko2022forming} considers a number of approaches for forming different predictive features of tweet data sets and using them in the predictive analysis for the decision-making support. The graph theory as well as frequent itemsets and association rules theory is used for forming and retrieving different features from these datasests. The use of these approaches makes it possible to reveal a semantic structure in tweets related to a specified entity. It is shown that quantitative characteristics of semantic frequent itemsets can be used in predictive regression models with specified target variables. Using  the graph theory, users’ communities and influencers can be revealed given tweets characteristics. 
In \cite{pavlyshenko2022methods},  different approaches for the analysis of news trends on Twitter have been considered.  
The obtained results show that an effective system for detecting fake and manipulative news can be developed using combined neural network which consists of three concatenated subnetworks. Discussions on social networks about companies' behavior have some impact on their business and their stock prices on the stock market. To analyze such an impact and make risk assessment, Bayesian regression can be used. Using the theory of frequent itemsets and association rules along with thematic fields of keywords, makes it possible to reveal the semantic structure for entities in news messages. 
LLMs are being effectively used in analysing  financial data and news. Paper~\cite{yang2023fingpt} 
considers an open-source large language model, FinGPT, for the finance sector.  FinGPT takes a data-centric approach, providing researchers and practitioners with accessible and transparent resources to develop their FinLLMs. 

In this study, we are going to consider and test the fine-tuning Llama 2 GPT large language model (LLM) model~\cite{touvron2023llama}  on financial dataset on multitask instructions,
such as  analysing a text from financial market perspectives, 
highlighting the main points of a text, summarizing a text and extracting named entities with appropriate sentiments. 

\section{Parameter-efficient Fine-Tuning Llama 2 LLM}
Llama 2 model is considered in the work~\cite{touvron2023llama}  as 
 a collection of pretrained and fine-tuned large language models (LLMs) ranging in scale from 7 billion to 70 billion parameters. Fine-tuned LLMs, called Llama 2-Chat, are optimized for dialogue use cases. These models outperform open-source chat models on most benchmarks they have been tested, and based on  human evaluations for helpfulness and safety, they may be a suitable substitute for closed-source models. The paper~\cite{touvron2023llama} provides a detailed description of an approach to fine-tuning and safety improvements of Llama 2-Chat. 
Full fine-tuning is applicable in the case when we need to ingest millions of documents into LLM. But in the case of much smaller data, we can use a PEFT/LoRA approach which consists in fine-tuning a much smaller number of model parameters. These parameters are saved in the model adapter which is used for full model modification before using it for the model text response generation.  To optimize GPU usage, 
4bit or 8bit quantization of LLM can be chosen for model fine-tuning.
State-of-the-art Parameter-Efficient Fine-Tuning (PEFT) methods enable efficient adaptation of pre-trained language models (PLMs) to various downstream applications without fine-tuning all the model's parameters. Fine-tuning large-scale PLMs is often prohibitively costly. In this regard, PEFT methods only fine-tune a small number of (extra) model parameters, thereby greatly decreasing the computational and storage costs. Recent State-of-the-Art PEFT techniques achieve the performance comparable to that of full fine-tuning~\cite{peft}.
The paper~\cite{hu2021lora} considers Low-Rank Adaptation, or LoRA, which freezes pre-trained model weights and injects trainable rank decomposition matrices into each layer of the Transformer architecture, greatly reducing the number of trainable parameters for downstream tasks.  PEFT/LoRA approach approach makes it possible to fine-tune LLMs with sizes near 7B parameters, using Google Colab.  Along with text data for fine-tuning, it is important to use prompt instructions which show how to process input prompts. Instructions can be created by human experts and augmented by other LLM models. 
LLM generate complex output texts on prompts which can be optimized by different ways. One possible way is selecting appropriate instructions for fine-tuning models. Another way is using a method called Reinforcement Learning from Human Feedback (RLHF)~\cite{beeching2023stackllama}. In this approach, human experts estimate and rate LLM output and then using this rates as target lables, LLM can be fine-tuned by supervised training.  

Let us consider fine-tuning Llama 2 using a PEFT/LoRA approach. 
The instruction for fine-tuning can be created in different ways, e.g. by experts or using LLMs like ChatGPT, GPT-4, Llama with appropriate prompts which specify LLM response.
 The dataset of financial news was taken from Kaggle repository~\cite{kagglefinnews}. It contains the data from the following news publishers: Bloomberg.com, CNBC.com, reuters.com, wsj.com, fortune.com. For testing the fine-tuning, we used dataset with prompt instructions for analysing a text, 
highlighting the main points of a text, summarizing a text and extracting named entities with appropriate sentiments.  The dataset was split into training and validation datasets, the relative part of validation dataset was 0.25.
The approaches for using PEFT/LoRA for fine-tuning Llama 2 are described in~\cite{hfllama2, vonwerra2022trl}.
 For the model fine-tuning the trainer \verb|SFTTrainer| 
from package \verb|trl|~\cite{vonwerra2022trl} was used.
The following training arguments were set up:
\begin{newmargin}{1cm}{0cm} 
\begin{verbatim}
'model_name':'meta-llama/Llama-2-7b-chat-hf',
learning_rate':5e-4, 
'num_train_epochs':10,
'max_seq_length':2048,
'gradient_accumulation_steps':2, 
'load_in_4bit': True for 4bit and False for 8bit model quantization 
'load_in_8bit': False for 4bit and True for 8bit model quantization
'bnb_4bit_quant_type':'nf4',
'lr_scheduler_type':'linear'
\end{verbatim}
\end{newmargin}

\begin{figure}[htb]
\center
\includegraphics[width=0.55\linewidth]{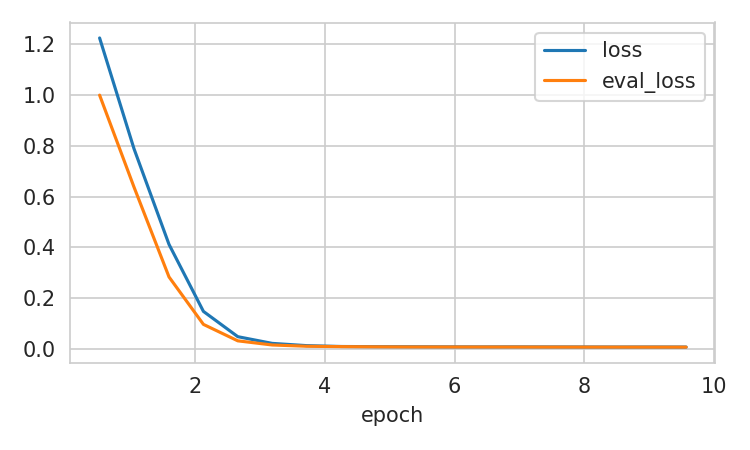}
\caption{Loss curves for training (\textit{loss}) and validation (\textit{eval\_loss}) datasets in case of 4bit model quantization.}
\label{fig1}
\end{figure}

\begin{figure}[htb]
\center
\includegraphics[width=0.55\linewidth]{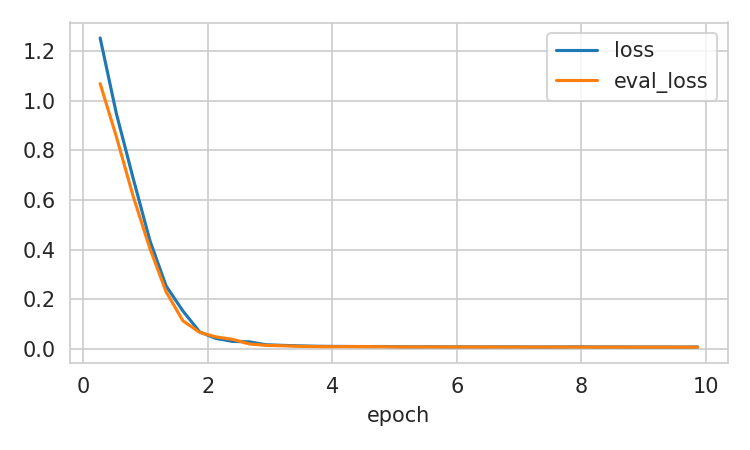}
\caption{Loss curves for training (\textit{loss}) and validation (\textit{eval\_loss}) datasets in case of 8bit model quantization.}
\label{fig2}
\end{figure}

Figures~\ref{fig1},~\ref{fig2} show the loss curves for training and validation datasets on epochs of LLM fine-tuning for 4bit and 8bit model quantization for fine-tuning respectively. One can see that loss curves are very similar for  4bit and 8bit model quantization. Which model generates more efficient 
responses can be  defined qualitatively by experts. 

\FloatBarrier

\section{Testing Fine-Tuned Llama 2 LLM}
For testing the performance of fine-tuned Llama 2 LLM, we have chosen several arbitrary news from the dataset~\cite{kagglefinnews}. 
The prompt for analysing the text using fine-tuned LLama 2 LLM has the following structure:
\begin{newmargin}{1cm}{0cm} 
\begin{verbatim}
<s>[INST] <<SYS>>
You are an expert in financial news analytics.
Please find companies, products, technologies and currencies 
in the text and assess sentiments towards them.
<</SYS>>
      
Please analyse the text:
{ Input text } [/INST]
\end{verbatim}
\end{newmargin}
where \verb|{ Input text }| -- is the text of financial news.

The fine-tuned LLM conducts the analysis of an input text with the following tasks: analysing the text from financial market perspectives, 
highlighting the main points of the text, summarizing the text and extracting named entities with appropriate sentiments. 
The examples of arbitrary input texts and LLM responses are as follows: 
\begin{newmargin}{1cm}{0cm} 
{\fontsize{9}{9} \fontfamily{lmtt} \selectfont text
 \textbf{Input text: }
 \\ The U.S. dollar just hit its highest levels since late December, a move that's planted it in overbought territory, according to one chart watcher. 
  "On a very short-term basis it's getting a little ahead of itself," Matt Maley , equity strategist at Miller Tabak, told CNBC's " Trading Nation " on Tuesday. He noted that the dollar's relative strength index, a measure of momentum, hit 76.07 on Wednesday, surpassing the one-year peak hit in October. 
  "It's getting to a level, actually slightly above a level, that it reached last fall just as the dollar was starting to finish off its little bit of a rally here and then rolled back over and made new lows," said Maley. 
  The DXY U.S. dollar index tried to buck a nine-month downward trend in late September, but its short burst of gains did not last through to the end of 2017. The greenback broke higher in April after struggling for direction through the start of this year. 
  Unlike last year's attempt to rally, the technicals suggest gains over the past month are a little different, said Maley. 
  "The rally has been much stronger. It took it above its 200-day moving," he said. "Also, above its trend line going all the way back to the highs in very early 2017. When you combine that with the kind of positioning we see where a lot of people are very short the dollar and very long currencies like the euro, I think that this kind of next pullback will be a little bit more shallow and could take the dollar higher." 
  The dollar touched its 200-day moving average on April 27 and broke above it on May 1. The index has not traded above that key level since May 2017. 
  U.S. economic growth and the expectation for higher interest rates should also give the rally in the dollar more fuel, said Gina Sanchez , CEO of Chantico Global. 
  "We're well into the recovery at this point and, in fact, Europe has largely peaked out, Japan has peaked out," Sanchez told "Trading Nation" on Tuesday. "However, the U.S. is continuing to expand and I think that that is very, very supportive of the expectation for higher interest rates." 
  The U.S. economy grew at a 2.3 percent pace in the first quarter of this year, level with full-year economic growth in 2017. Japan's economy expanded 1.7 percent in 2017, while euro zone GDP increased 2.5 percent. 
  "We're actually seeing inflation come in at Fed-expected levels whereas European inflation is starting to disappoint," said Sanchez. "That's also going to be supportive of a higher dollar from here." 
  Higher inflation expectations in the U.S. have prompted more hawkish talk from the Federal Reserve . Markets anticipate at least two more interest rate hikes this year after an increase in March, according to CME Group fed funds futures. A fourth increase in December remains a possibility. 
  Vote Vote to see results Total Votes: Not a Scientific Survey. Results may not total 100\% due to rounding. 
  Disclaimer {[}/INST{]} 
 \\  
 \\ \textbf{Response:} \\ 
 \\  Analysis: 
 \\ The text describes recent financial performances of the U.S. dollar. It has reached its highest levels since late December. Matt Maley, an equity strategist at Miller Tabak, argues that this development leads the dollar into overbought territory. He points out that the dollar's relative strength index (a measure of momentum) has surpassed last year’s peak. However, Maley suugests that the rally in the dollar is stronger than last year's and rise could continue if the U.S. economy continues to expand given the positive economic climate in the country. Gina Sanchez, CEO of Chantico Global, seems to agree with this viewpoint, stating that the U.S.'s economic growth and anticipation of increasing interest rates will likely support a sustained dollar rally. 
 \\  
 \\ Main Points: 
 \\ 1. The U.S. dollar has hit high levels, leading it into overbought territory.  
 \\ 2. It surpasses the relative strength index one-year peak. 
 \\ 3. Unlike last year's attempt, current gains are stronger, suggesting the dollar might go higher. 
 \\ 4. The U.S.'s robust economic growth and the possibility of higher interest rates support the dollar rally. 
 \\ 5. Expected inflation levels in the U.S. accompany more hawkish talk from the Federal Reserve. 
 \\  
 \\ Summary: 
 \\ The U.S. dollar has hit its highest level since late December, indicating it might be overbought; still, the rise might continue due to encouraging U.S. economic growth and speculation on higher interest rates. 
 \\  
 \\ JSON Data: 
 \\ {[}\{"entity":"U.S. dollar", "entity\_name":"currency", "sentiment":"positive"\}, \{"entity":"Matt Maley", "entity\_name":"person", "sentiment":"neutral"\}, \{"entity":"Miller Tabak", "entity\_name":"company", "sentiment":"neutral"\}, \{"entity":"Gina Sanchez", "entity\_name":"person", "sentiment":"neutral"\}, \{"entity":"Chantico Global", "entity\_name":"company", "sentiment":"neutral"\}, \{"entity":"Federal Reserve", "entity\_name":"organization", "sentiment":"neutral"\}, \{"entity":"U.S. economy", "entity\_name":"economy", "sentiment":"positive"\}, \{"entity":"interest rates", "entity\_name":"financial product", "sentiment":"neutral"\}{]}   
 \\  
 \\
 ------------------------------------------------------------------------------------
 \\
 \\ Input text: 
 \\ Toy maker Hasbro has agreed to acquire children's entertainment and merchandising franchises, including the characters of the superhero TV show Power Rangers, from Saban Properties in a deal valued at \$522 million in cash and stock, the companies announced Tuesday. 
  The deal comes as Hasbro, which is the world's largest toy maker with a stable of franchises including "My Little Pony," "Monopoly" and "The Transformers," seeks to reverse its losses following the bankruptcy last year of U.S. toy retailer Toys R Us. 
  In addition to making toys and action figures, Hasbro profits from such franchises through the production of movies and TV series, allowing it to diversify its revenue beyond retail sales. 
  The deal with Saban Properties was first reported by Reuters earlier on Tuesday. 
 Under the agreement, Hasbro will launch its first set of products off the Saban franchises, which include Luna Petunia, Julius Jr and Popples, in 2019. 
  Launched as the Mighty Morphin Power Rangers live-action TV show in 1993, the franchise was created by Haim Saban, owner of Saban Properties. 
  The TV series gave rise to a line of action figures and other merchandise, plus three movies, including "Sabans Power Rangers" last year. The 2017 film, distributed by Lions Gate Entertainment , sold \$142 million worth of tickets worldwide, according to Box Office Mojo. 
  Earlier this year, Hasbro signed a deal with Saban for the toy maker to design, produce and bring to market a wide variety of toys and role-play items inspired by Power Rangers. 
  Pawtucket, Rhode Island -based Hasbro has taken several steps to boost its presence in the entertainment business as a way to fuel toy sales. 
  The company operates Hasbro Studios, which produces TV shows such as the Netflix series, Stretch Armstrong and the Flex Fighters. 
  The toy industry's traditional players have been undone in recent years by a shift toward thousands of rival, smaller producers selling on Amazon and other e-commerce sites, as well as kids' preference for electronic games over physical toys. 
  Last week, Hasbro reported a net loss attributable to the company of \$112.5 million, or 90 cents per share, in the first quarter ended April 1, compared with a profit of \$68.6 million, or 54 cents per share, a year earlier. 
  In 2014, Hasbro held merger discussions with DreamWorks Animation, the studio behind Shrek, but DreamWorks was subsequently bought by Comcast . 
  Last year, the toymaker also held talks to acquire U.S. movie studio and entertainment company Lions Gate, but those negotiations broke down over price, sources said at the time. 
  Hasbro has been seeking scale, and has attempted unsuccessfully to merge with peer Mattel over the years, most recently in 2017. Adding to the sense of crisis in the toy sector, Mattel appointed its fourth chief executive in three years last week. 
  J.P. Morgan Securities served as the financial adviser to Hasbro, according to the announcement. 
  Disclosure: Comcast is the parent company of CNBC. {[}/INST{]} 
 \\  
 \\ \textbf{Response:} 
 \\  
 \\  Analysis: 
 \\ The text discusses the acquisition of Saban Properties' franchises by world-renowned toy manufacturer Hasbro. This move comes after Hasbro suffered losses following the bankruptcy of Toys R Us. Hasbro's strategy implies entering the entertainment sectors to stimulate toy sales. However, the firm's financial performance has been weaker recently, with the first 2018 quarter ending with a loss. Hasbro's previous attempts to merge with DreamWorks Animation and Lions Gate were unsuccessful, and it is also facing competition from smaller producers selling on e-commerce platforms.  
 \\  
 \\ Main Points: 
 \\ 1. Hasbro agrees to acquire merchandising franchises, including Power Rangers from Saban Properties. 
 \\ 2. The deal, worth \$522 million, follows Hasbro's losses after the bankruptcy of Toys R Us. 
 \\ 3. Hasbro plans to diversify revenue sources by entering the entertainment sector. 
 \\ 4. The company's financial difficulties are clear as it reported a significant loss in Q1 2018. 
 \\ 5. Hasbro's efforts to merge with other companies like DreamWorks Animation and Lions Gate Entertainment have been unsuccessful. 
 \\  
 \\ Summary: 
 \\ Hasbro Inc. engages in a deal to acquire franchises from Saban Properties LLC in a bid to reverse losses after Toys R Us' bankruptcy and to diversify income sources towards entertainment business, despite facing a significant financial slump. 
 \\  
 \\ JSON Data: 
 \\ {[}\{"entity":"Hasbro", "entity\_name":"company", "sentiment":"negative"\}, \\ \{"entity":"Saban Properties", "entity\_name":"company", "sentiment":"neutral"\}, \\ \{"entity":"Toys R Us", "entity\_name":"company", "sentiment":"negative"\}, \\ \{"entity":"Power Rangers", "entity\_name":"product", "sentiment":"positive"\}, \\ \{"entity":"Lions Gate Entertainment", "entity\_name":"company", "sentiment":"neutral"\}, \\ \{"entity":"Amazon", "entity\_name":"company", "sentiment":"positive"\}{]}   
 \\  
 \\
 ------------------------------------------------------------------------------------
 \\  
 \\ Input text: 
 \\ UPDATE 1-Trump tells farmers he supports crop insurance Jeff Mason Reuters SHARES  
  (Recasts to add farm bill and crop insurance, changes to show Trump signed executive order, adds reaction)  
  NASHVILLE, Jan 8 (Reuters) - U.S. President Donald Trump on Monday told farmers gathered in Tennessee that he supports crop insurance, signaling a possible shift from his proposals last year to shrink a program that many farmers rely on for funding their operations.  
  Speaking at the American Farm Bureau's annual convention, Trump said Congress would include the subsidies for crop insurance premiums in the upcoming "farm bill" that directs billions of dollars each year to agriculture, nutrition, food and land-conservation programs.  
  Im looking forward to working with Congress to pass the farm bill on time so that it delivers for all of you and I support a bill that includes crop insurance," Trump said after noting Senate Agriculture Committee Chairman Pat Roberts was in attendance. Were working hard on the farm bill and I think its going to go well.  
  In May Trump angered farm groups, which represent the rural voters who generally support his Republican Party, when he proposed cutting \$46.54 billion in federal agricultural funding, mostly by capping the amount the government chips in toward crop insurance. There currently is no limit to the support the government gives farmers for insurance premiums.  
  Trump, who swept Tennessee in the 2016 election, also signed an executive order after the speech to make it easier for the private sector to locate broadband infrastructure on federal land and buildings, part of a push to expand high-speed internet in rural America. Faster internet speeds in rural areas are often considered key to addressing the economic divide between rural and urban America, but the costs have so far been prohibitive.  
  About 39 percent of rural Americans lack access to high- speed internet service, the Federal Communications Commission said in a 2016 report.  
  "We need to get rural America more connected. We need it for our tractors, we need it for our schools, we need it for our home-based businesses," a White House official told reporters ahead of Trump's speech.  
  The White House described the moves as an incremental step to help spur private development while the administration figures out how it can help with funding, something that could become part of Trump's plan to invest in infrastructure.  
  The CEO of the leading telecommunications trade group in the country, Jonathan Spalter, said USTelecom would push to ensure the government sets aside money for helping build and upgrade network infrastructure, as well as dismantling and streamlining regulations. (Additional reporting by Eric Beech, Lisa Lambert and Roberta Rampton in Washington; Editing by Lisa Shumaker) {[}/INST{]} 
 \\  
 \\ \textbf{Response:} 
 \\  
 \\  Analysis:  
 \\ The text discusses recent actions from U.S. President Donald Trump, aimed at supporting farmers and advancing rural internet connectivity. Regarding crop insurance, Trump is indicating a departure from his previous plans to scale back a program widely used by farmers for operational funding. His new stance is seen as favourable towards the agribusiness sector and particularly for crop insurance companies, signaling potential growth in their market due to official support. Trump's push for expanded broadband service in rural areas indicates opportunities for telecommunications companies, particularly those involved in infrastructure building and upgrades. His plans for streamlining of regulations could also provide a positive environment for these firms. However, the absence of concrete funding plans from the administration injects a note of uncertainty into the picture. 
 \\  
 \\ Main Points: 
 \\ 1. President Trump expressed his support for crop insurance subsidies, contrasting with his prior proposals to reduce federal agricultural funding. 
 \\ 2. Trump signed an executive order to facilitate the placement of private-sector broadband infrastructure on federal land, aiming to enhance internet speed in rural America. 
 \\ 3. About 39 percent of rural Americans still lack access to high-speed internet service, a gap the administration aims to close. 
 \\ 4. The White House characterized these actions as initial steps toward incentivising private sector investment while defining its funding strategies. 
 \\ 5. USTelecom and its CEO, Jonathan Spalter, aim to lobby the government to allocate resources for network infrastructure development and regulation overhaul. 
 \\  
 \\ Summary: 
 \\ Trump's pivot towards support for crop insurance and his push for rural internet infrastructure indicate potential opportunities for agriculture and telecom companies, despite uncertainty about funding strategies.  
 \\  
 \\ JSON Data: 
 \\ {[} 
 \\   \{"entity":"crop insurance", "entity\_name":"product", "sentiment":"positive"\}, 
 \\   \{"entity":"Trump", "entity\_name":"company", "sentiment":"positive"\}, 
 \\   \{"entity":"agricultural funding", "entity\_name":"product", "sentiment":"negative"\}, 
 \\   \{"entity":"rural America", "entity\_name":"group", "sentiment":"positive"\}, 
 \\   \{"entity":"broadband infrastructure", "entity\_name":"product", "sentiment":"positive"\}, 
 \\   \{"entity":"USTelecom", "entity\_name":"company", "sentiment":"positive"\}, 
 \\   \{"entity":"Jonathan Spalter", "entity\_name":"person", "sentiment":"neutral"\} 
 \\ {]}    
 \\  
 \\
 ------------------------------------------------------------------------------------
 \\
 \\ Input text:  
 \\ Tesla is apparently still deep in the circles of production hell. 
  On Wednesday, the electric car maker released delivery numbers for the fourth quarter of 2017 that fell short of many expectations on Wall Street, and once again pushed production targets on its highly anticipated Model 3 sedan.  
  Tesla shares fell roughly 2 percent in after-hours trading. 
  "As we continue to focus on quality and efficiency rather than simply pushing for the highest possible volume in the shortest period of time, we expect to have a slightly more gradual ramp through Q1, likely ending the quarter at a weekly rate of about 2,500 Model 3 vehicles," Tesla said in a release. "We intend to achieve the 5,000 per week milestone by the end of Q2." 
  In 2017, the company had said it planned to reach a production rate of 5,000 cars per week for the Model 3, but later revised back that target to the end of the first quarter. Now, Tesla expects to reach the target by the end of the second quarter. 
  Tesla said it delivered 29,870 vehicles in the fourth quarter of 2017, including 1,550 of its anticipated Model 3 sedan.  
  The California-based electric-car maker also delivered 15,200 Model S sedans, and 13,120 Model X SUVs. 
  That compares with 22,200 vehicles in the fourth quarter of 2016, of which 12,700 were Model S and 9,500 were Model X. Total 2016 deliveries were roughly 76,230. There were no Model 3 deliveries in 2016, as the car was not in production yet. 
  In addition to those deliveries, Tesla said there were 2,520 Model S and X vehicles and 860 Model 3 vehicles in transit to customers at the end of the quarter, which Tesla will count as deliveries in Q1 2018.  
  Tesla said it made "major progress" toward addressing the "production bottlenecks" the company blamed for falling so far short of its Model 3 targets in the third quarter.  
  Tesla had said at the end of the third quarter it would deliver about 100,000 Model S and X vehicles in 2017, a 31 percent increase over the previous year. CEO Elon Musk said he expected weekly Model 3 production to be "in the thousands" by the end of 2017.  
  Wall Street estimates varied widely. In a note sent Tuesday, Cowen analyst Jeffrey Osborne estimated a Wall Street consensus 4,000-5,000 Model 3 deliveries in the fourth quarter . But Osborne's own forecast was below that at just 2,250. Oppenheimer analyst Colin Rusch had expected Tesla to deliver just 800 Model 3 cars. 
 "In the 12-month investable time frame our rating contemplates, we see Tesla as a great company led by a true visionary, but must acknowledge the asymmetric risk/reward profile for the stock at the market's current valuation," Osborne said in the note. "Simply, we see a lot more that can go wrong than can go right as the company transitions into Mr. Musk's greater vision as laid out in his Master Plan, Part Deux." {[}/INST{]} 
 \\  
 \\ \textbf{Response:} 
 \\  
 \\  Analysis: 
 \\ Tesla has faced some turbulence in its vehicle production, particularly its Model 3 sedan, with delivery numbers for Q4 2017 coming in below Wall Street expectations. This shortfall in performance resulted in a drop in Tesla shares after-hours by approximately 2\%. Despite this, Tesla affirmed their focus on quality and efficiency over quantity, with plans to gradually increase Model 3 production through to Q2 2018. Despite the marked setbacks, Tesla delivered a significant volume of vehicles in Q4 2017 and said it made "major progress" toward addressing production issues. Analyst opinions on Tesla’s performance vary widely, although some express optimism for Tesla’s vision but caution over the stock’s risk/reward profile. 
 \\  
 \\ Main Points: 
 \\ 1. Tesla’s Q4 2017 production delivery numbers for the Model 3 fell short of Wall Street expectations. 
 \\ 2. Tesla shares fell by approximately 2\% in after-hours trading. 
 \\ 3. Despite setbacks, Tesla delivered 29,870 vehicles in Q4 2017. 
 \\ 4. Analyst's view on Tesla's performance vary, with some expressing cautious optimism. 
 \\ 5. Tesla plans to gradually increase Model 3 production through to Q2 2018, focusing on quality and efficiency over volume. 
 \\  
 \\ Summary: 
 \\ Tesla’s output of the anticipated Model 3 sedan fell short of expectations, leading to a share value drop, yet the company remains optimistic and plans to increase production. 
 \\  
 \\ JSON Data: 
 \\ {[}\{"entity": "Tesla", "entity\_name": "company", "sentiment": "negative"\}, \\ \{"entity": "Model 3 sedan", "entity\_name": "product", "sentiment": "negative"\}, \\ \{"entity": "Tesla shares", "entity\_name": "company", "sentiment": "negative"\}, \\ \{"entity": "Model 3 vehicles", "entity\_name": "product", "sentiment": "neutral"\}, \\ \{"entity": "Elon Musk", "entity\_name": "company", "sentiment": "neutral"\}, \\ \{"entity": "Cowen", "entity\_name": "company", "sentiment": "neutral"\}, \\ \{"entity": "Oppenheimer", "entity\_name": "company", "sentiment": "neutral"\}{]}   
 \\  
 \\
 ------------------------------------------------------------------------------------
 \\ 
 \\ Input text:  
 \\ As the dollar charges to fresh 2018 highs, one trader is betting on trouble ahead for the emerging markets. 
  Global stocks have been a hot trade for investors with the iShares MSCI emerging markets ETF (EEM) surging more than 15 percent in the past year, outperforming the S\&P 500, which is up just 10 percent in that time. However, Dan Nathan, co-founder and editor of RiskReversal.com, warns the options market could be predicting that the Chinese tech-heavy ETF could be in jeopardy. 
  Tencent , Samsung, Alibaba and Taiwan Semiconductor make up roughly 16 percent of the EEM. While the stocks have been hot over the last year, they are all down sharply from their recent highs. 
  Tuesday on CNBC's " Fast Money," Nathan noted that there was an abundance of bearish activity this week. 
  Specifically, Nathan pointed out a trader's purchase of 21,000 June 42/45 put spreads, paying 55 cents per contract. This is a bearish bet that the emerging markets ETF will fall below \$44.45 by June, or down nearly 5 percent from its current levels. 
  "Pretty interesting breakdown level in \$42, {[}it{]} may be some decent support surrounding a 'toppy' sort of feel" he said. "So maybe this is a little protection against a long position in the EEM." 
  Shares of the emerging markets ETF have fallen 1 percent so far this year and were down Wednesday afternoon at around \$46.60. {[}/INST{]} 
 \\  
 \\ \textbf{Response:}   
 \\  Analysis: 
 \\ The text revolves around the rise in the value of the dollar, the performance of the iShares MSCI emerging markets ETF (EEM), and the impact on several companies. The sentiment towards the dollar is positive as it reaches fresh 2018 highs. The narrative also sheds a positive light on the global stocks and EEM, which outperformed S\&P 500 by over 15\% in the past year. However, sentiments change to negative regarding EEM as options market predicts potential future jeopardy for this ETF. Tencent, Samsung, Alibaba and Taiwan Semiconductor collectively make up 16\% of EEM, and they were all portrayed negatively as they have fallen sharply from their recent highs. Despite EEM's past performance, the platform of RiskReversal.com anticipates a 5\% downfall in the share value of EEM by June. The EEM has already faced a 1\% decrease this year. 
 \\  
 \\ Main Points: 
 \\ 1. The value of the dollar is reaching new highs in 2018. 
 \\ 2. The iShares MSCI emerging markets ETF (EEM) has outperformed the S\&P 500 by over 15\% in the past year. 
 \\ 3. The Chinese tech-heavy ETF could be in jeopardy as predicted by the options market. 
 \\ 4. Companies like Tencent, Samsung, Alibaba, and Taiwan Semiconductor, have lost significant value from their recent highs. 
 \\ 5. RiskReversal.com predicts about a 5\% decrease in EEM's value by June, which is already evident with a 1\% decrease this year. 
 \\  
 \\ Summary: 
 \\ The iShares MSCI emerging markets ETF (EEM) could face potential downslide as the dollar strengthens and certain tech companies lose their value. 
 \\  
 \\ JSON Data: 
 \\ {[}\{"entity":"Dollar","entity\_name":"currency","sentiment":"positive"\}, \\ \{"entity":"global stocks","entity\_name":"product","sentiment":"positive"\}, \\ \{"entity":"iShares MSCI emerging markets ETF (EEM)",\\  "entity\_name":"product","sentiment":"negative"\}, \\ \{"entity":"Tencent","entity\_name":"company","sentiment":"negative"\},  \{"entity":"Samsung","entity\_name":"company","sentiment":"negative"\}, \\  \{"entity":"Alibaba","entity\_name":"company","sentiment":"negative"\}, \\ \{"entity":"Taiwan Semiconductor","entity\_name":"company","sentiment":"negative"\}{]} 
}
\end{newmargin}
\section{Conclusion}
 The main goal of the study is to see if the analytical capability of the fine-tuned large language model has a predictive potential and can generate a response with a specified structure which can be useful in financial analytics.
In the study, we have considered the possibility to fine-tune Llama 2 GPT large language model (LLM) for multitask analyses of financial news. For fine-tuning, the PEFT/LoRA based approach was used. The model was fine-tuned for the following tasks: analysing the text from financial market perspectives, 
highlighting the main points of the text, summarizing the text and extracting named entities with appropriate sentiments. The PEFT/LoRA approach makes it possible to use cheap GPU resources for model fine-tuning. 
The obtained results show that fine-tuned Llama 2 model can perform multitask financial news analyses with a specified structure of response, part of response can be a structured text and another part of data can have a JSON format that is convenient for further processing LLM response. 
The PEFT/LoRA approach makes it possible to use cheap GPU resources for model fine-tuning which are available, e.g. on Google Colab  in case of small input text size and 8bit or 4bit quantization.
In the test results, one can mention some inaccuracy, e.g. in specifying names of entities in JSON output. This problem can be fixed using more accurate and specified training dataset with the training dataset with instructions corrected  by an expert. 

The multitask response of the model can be used by financial analysts who analyse large amount of news data. Taking into account that LLM can generate output in specified JSON format, the data of sentiments for named entities can be used in predictive models as features and can be loaded directly into predictive models via appropriate API. 
 These features can have a predictive potential for different target variables including quantitative characteristics 
 of companies' behavior on the financial markets. Let us mention some  possible predictive analytic  approaches where sentiments of named entities can be used as predictive features. 
In the survey~\cite{pavlyshenko2022analytics}  we consider the case studies on sales time series forecasting, the deep learning approach for forecasting non-stationary time series using time trend correction, dynamic price and supply optimization using Q-learning. In~\cite{pavlyshenko2020using}, we study Bayesian regression for building time series models and stacking different predictive models for time series. the use of Bayesian regression for time series modeling with nonlinear trend was analyzed. This approach makes it possible to estimate an uncertainty of time series prediction and calculate value at risk characteristics.  The probabilistic approach for stacking predictive models allows us to make risk assessment for the predictions that are important in a decision-making process.
In~\cite{pavlyshenko2020sales}, we study the use of deep Q-learning models in the problems of sales time series analytics. In contrast to supervised machine learning which is a kind of passive learning using historical data, Q-learning is a kind of active learning aimed at maximizing a reward by an optimal sequence of actions of an intellectual agent. The sequence of actions of intellectual agent can be optimized by maximizing the reward function when the environment for learning agent interaction can be modeled using the parametric model and in the case of using the model which is based on the historical data. 
 In the above mentioned predictive analytics approaches, the extracted sentiments from texts using fine-tuned LLM can be considered as predictive categorical features with possible values \textit{'negative', 'neutral', 'positive'}. The predictive potential of these sentiment features can be studied using the cross-validation approach for predictive models. The use of the Bayesian approach for stacking predictive models allows us to make different kinds of risk assessments to support decision making. E.g. having the probability density function for target variable, one can calculate Value at Risk (VaR) as 0.05 quantile for the target variable. 

The obtained results on fine-tuning  Llama2 GPT large language model (LLM) using PEFT/LoRA approach can be assessed  qualitatively by experts to see if the news analytics, summarizing, highlighting main points and entities extracting give some essential information to market experts.  
The considered approach can show high efficiency, using small sets of instructions due to the LLM ability of few-shot learning that is not inherent for conventional transformer based models. To improve the LLM performance, one needs a more precisely created training dataset and to exploit the RLHF method to get a more optimized LLM. The considered approach can be applied for using extracted entities and sentiments in supervised models with quantitative target variables, e.g. for the analysis of companies' behavior on financial and business markets. 

\section{Disclaimer} 
We are sharing a considered approach, ideas and results for academic purpose only, not any financial advice or recommendations for real business or investment.

\bibliographystyle{unsrt}
\bibliography{article.bib}
\end{document}